\documentclass[12pt]{article}

\usepackage[utf8]{inputenc}
\usepackage[T1]{fontenc}
\usepackage[english]{babel}
\usepackage[top=1in, bottom=1in, left=1in, right=1in]{geometry}
\usepackage{setspace}
\usepackage{fancyhdr}
\newcommand{\runningheads}[2]{
    \setlength{\headheight}{15pt}
    \addtolength{\topmargin}{-2pt}
    \pagestyle{fancy}  % First page forced plain, 2nd+ fancy 
    \fancyhead{}  % Reset all header and footer  
    \fancyhead[L]{\small #1}
    \fancyhead[R]{\small #2}
    \renewcommand{\headrulewidth}{0.4pt}
}

\usepackage{float}
\usepackage{graphicx}
\usepackage{subcaption}
\usepackage{booktabs}
\usepackage{makecell}
\usepackage{multirow}
\usepackage{threeparttable}  % for table footnotes

\usepackage[linesnumbered,ruled,vlined]{algorithm2e}
\SetCommentSty{textnormal}
\usepackage{enumitem}
\usepackage{pdfpages}

\usepackage{amsmath}
\usepackage{amssymb}
\usepackage{amsthm}
\DeclareMathOperator*{\argmax}{arg\,max}

\theoremstyle{definition}

\usepackage{authblk}

\usepackage{natbib}
\usepackage{xcolor}
\definecolor{darkblue}{rgb}{0, 0, 0.75}
\definecolor{darkgreen}{rgb}{0, 0.75, 0}
\usepackage[colorlinks=true, linkcolor=black, citecolor=darkgreen, urlcolor=darkblue]{hyperref}

\newenvironment{foldable}{}{}
\newcommand*{\ie}{i.e.}
\newcommand*{\eg}{e.g.}
\newcommand*{\etc}{etc.}
\newcommand{\meanstd}[2]{#1$_{\pm{#2}}$}
\newcommand{\meanstdbf}[2]{\textbf{#1}$_{\pm{#2}}$}
\newcommand{\whitebox}{\textsuperscript{(W)}}
\newcommand{\blackbox}{\textsuperscript{(B)}}
\newcommand{\dg}{$^\dagger$}

\title{Improving Diversity in Black-box Few-shot Knowledge Distillation}
\author[1]{Tri-Nhan Vo}
\author[1]{Dang Nguyen}
\author[1]{Kien Do}
\author[1]{Sunil Gupta}
\affil[1]{Applied Artificial Intelligence Institute (A2I2), Deakin University, Australia}
\affil[1]{\texttt{\{s223032975, d.nguyen, k.do, sunil.gupta\}@deakin.edu.au}}
\runningheads{T-N. Vo et al.}{Improving Diversity in Black-box Few-shot KD}
\date{\small{Accepted at ECML-PKDD 2024}}

\begin{document}

\maketitle

\begin{abstract} \label{sec:00-abstract}
  Knowledge distillation (KD) is a well-known technique to effectively compress a large network (teacher) to a smaller network (student) with little sacrifice in performance.
However, most KD methods require a large training set and internal access to the teacher, which are rarely available due to various restrictions.
These challenges have originated a more practical setting known as \textit{black-box few-shot KD}, where the student is trained with few images and a black-box teacher.
Recent approaches typically generate additional synthetic images but lack an active strategy to promote their \textit{diversity}, a crucial factor for student learning.
To address these problems, we propose a novel training scheme for generative adversarial networks, where we adaptively select \textit{high-confidence} images under the teacher's supervision and introduce them to the adversarial learning on-the-fly.
Our approach helps expand and improve the diversity of the distillation set, significantly boosting student accuracy.
Through extensive experiments, we achieve state-of-the-art results among other few-shot KD methods on seven image datasets.
The code is available at \url{https://github.com/votrinhan88/divbfkd}.

\end{abstract}

\section{Introduction} \label{sec:01-introduction}
\begin{foldable}
  Over the recent years, deep learning models have made impressive progress---from classifying millions of images~\cite{russakovsky2015imagenet} to intelligently conversing with humans on various modalities and tasks.
  However, their ever-increasing sizes are a constant barrier to real-world deployment.
  This upscaling trend is the primary concern of model compression techniques, which aim to deliver powerful models at a deployable size (\eg, for mobile phones or embedded systems).
  While the removal of redundancy, quantization, or parameter sharing in the original model can help fulfill this goal, \textit{knowledge distillation} (KD)~\cite{hinton2015distilling} stands out for its efficiency and versatility.
\end{foldable}

\begin{foldable}
  The main idea of KD is to transfer the knowledge from a complex pre-trained network (\textit{teacher}) to a simpler network (\textit{student}).
  The pioneering solution was to train the student to simultaneously give correct predictions and mimic the teacher's outputs~\cite{hinton2015distilling}.
  With additional guidance, the student can approximate the teacher's performance despite its smaller size.
  This work has brought forth a new research area that most current KD methods are built upon~\cite{chen2017learning,meng2019conditional,nguyen2021knowledge}.
\end{foldable}

\begin{foldable}
  The fruitful performance of KD typically relies on the availability of an extensive training set to train the student.
  Traditional KD methods often assume that the student is distilled with the same data used for training the teacher~\cite{hinton2015distilling,kim2018paraphrasing,ahn2019variational,tian2020contrastive}.
  However, in real-world scenarios, the distillation often happens at an external party side, where one can only afford a limited-size training set.
  For example, to distill Google's FaceNet (trained using approximately 200 million non-public face images~\cite{schroff2015facenet}) into an in-house model without access to a large-scale dataset like Google's, an external party may only be able to gather a few thousand images or even less.
  This scenario poses the \textit{few-shot} setting and makes it more challenging than standard KD.
\end{foldable}

\begin{figure}
  \centering
  \includegraphics[width=0.99\columnwidth]{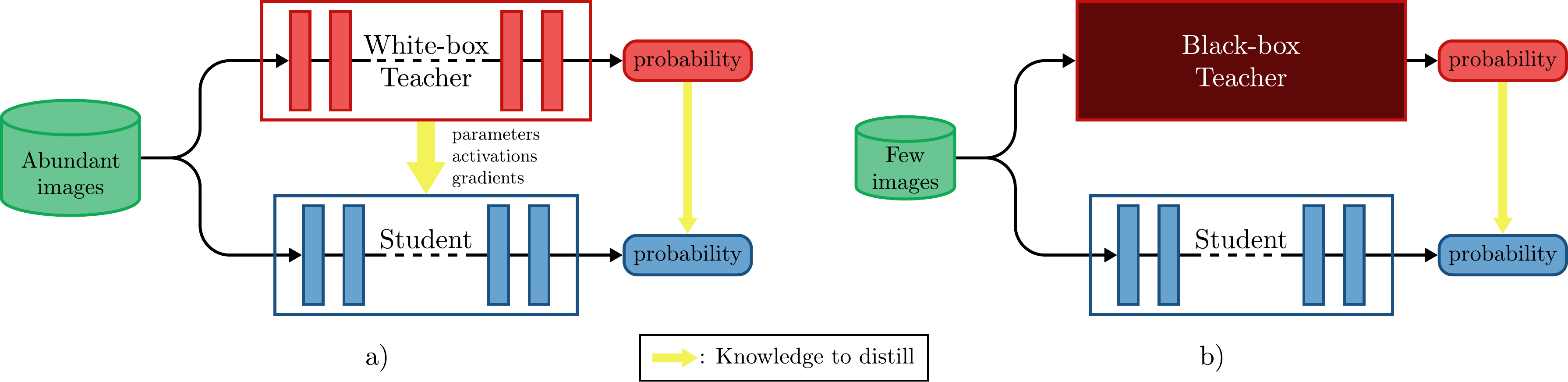}
  \caption{
    Common KD methods (a) assume no constraints whereas black-box few-shot KD, and (b) only has access to a few images and the teacher's predictive probabilities.
  }
  \label{fig:01-introduction-standard_kd_vs_bbfskd}
\end{figure}

\begin{foldable}
  Another assumption made by traditional KD methods is the requirement of a \textit{white-box} teacher \ie, we have complete access to its parameters, activations, or gradients, \etc~\cite{romero2014fitnets,yim2017gift,kimura2018few,chen2019data}.
  However, the teacher model is often \textit{black-box}, \ie, only its final predictions are accessible.
  Understandably, companies often avoid full disclosure of the model parameters as it is their competitive edge and intellectual property.
  For instance, even with a paid subscription to ChatGPT API~\cite{openai2022}, only text outputs and their prediction confidence can be retrieved.
  The realistic constraints in the \textit{black-box few-shot} KD setting (Fig.~\ref{fig:01-introduction-standard_kd_vs_bbfskd}) render most KD methods inapplicable and demand novel approaches.
\end{foldable}

\begin{foldable}
  Recently, some methods have addressed the few-shot KD problem, but most of them still require a white-box teacher~\cite{kimura2018few,kong2020learning}.
  To the best of our knowledge, only two methods apply a black-box teacher to few-shot KD, namely BBKD~\cite{wang2020neural} and FS-BBT~\cite{nguyen2022black}.
  Both these methods leverage MixUp~\cite{zhang2018mixup} to generate synthetic images.
  However, as MixUp linearly interpolates between two original images, the synthetic images are either very similar to the original images or do not have realistic semantics, adding little value to the student's training.
  To overcome this problem, FS-BBT employs a conditional variational autoencoder (CVAE)~\cite{sohn2015learning} to generate out-of-distribution synthetic images.
  However, while the pixel-based loss in CVAE is beneficial for reconstructing images, it is not intended for promoting image diversity and have side-effect blurring artifacts~\cite{bond2021deep}.
\end{foldable}

\begin{foldable}
  We address the problem of black-box few-shot KD \ie, distilling the student with only a few images and a black-box teacher.
  To boost student performance, we expand the distillation set with diverse synthetic images.
  To achieve diversity in the synthetic images, we propose a novel training scheme for Wasserstein Generative Adversarial Network (WGAN)~\cite{arjovsky2017wasserstein}:
  \begin{itemize}
    \item {
        Firstly, we specifically define \textit{high-confidence} images, which are synthetic images that the teacher can predict confidently w.r.t.\ a threshold.
        To avoid biased and abnormal images, we leverage the teacher to compute \textit{adaptive thresholds} in a class-specific and dataset-agnostic manner.
      }
    \item {
        Then, during the WGAN training loop, we select high-confidence images from the latest iteration on-the-fly and introduce them in the role of additional real images.
        Our WGAN learned with high-confidence images can generate diverse synthetic images close to the teacher's training images.
      }
    \item Finally, we augment our few-shot images with synthetic images to improve the training of the student, leading to a significantly better performance.
  \end{itemize}
\end{foldable}

\begin{foldable}
  We summarize our contributions as follows:
  \begin{enumerate}
    \item We propose \textbf{Diverse Black-box Few-Shot Knowledge Distillation} (\textbf{DivBFKD}), a novel solution for black-box few-shot KD, which directly addresses the diversity aspect of synthetic images to improve distillation.
    \item We propose a novel training scheme for WGAN, introducing high-confidence images efficiently and on-the-fly to the adversarial learning, which improves diversity in image generation.
    \item {
        We propose adaptive thresholds, a teacher-based criteria to avoid selecting biased and abnormal high-confidence images.
        The criteria can be adaptively applied to arbitrary classes and datasets.
      }
    \item {
        We conduct comprehensive evaluation on multiple image classification tasks and architectures, achieving SOTA results among few-shot KD methods.
        We also provide thorough analysis of our method for further insights.
      }
  \end{enumerate}
\end{foldable}

\section{Related Works} \label{sec:02-related}
\subsection{Knowledge Distillation}
\begin{foldable}
  Transferring knowledge from a large model (teacher) to a smaller model (student) was first introduced in~\cite{bucila2006model}, then formally popularized in~\cite{hinton2015distilling}.
  Its key idea is to exploit the knowledge encoded in the teacher's `softened' probabilistic outputs, which are continuous and often more informative than the `hard' class labels.
  The additional supervision helps the student capture underlying inter-class relationships and improve its performance.
\end{foldable}

\begin{foldable}
  Existing KD methods often involve logits-based, feature-based, and relation-based knowledge~\cite{gou2021knowledge}.
  Logits-based methods use the teacher's logits (\ie, pre-softmax activations) as the guidance for the student~\cite{ba2014deep,hinton2015distilling,nguyen2021knowledge}, optionally with adaptive temperature~\cite{li2023curriculum} or an ensemble of teachers~\cite{guo2020online}.
  Feature-based methods extract knowledge from the teacher's internal attributes \eg, parameters and inter-layer connections~\cite{liu2019knowledge}, intermediate features~\cite{romero2014fitnets}, attention maps~\cite{zagoruyko2016paying}, and feature-space probability distribution~\cite{passalis2018learning}.
  Finally, relation-based methods distill the knowledge in the relationships between different layers or data samples~\cite{park2019relational,yim2017gift,chen2020learning}.
  Although useful, these methods would struggle when only a few images are available in the distillation set and we can only access the black-box teacher's final predictions.
\end{foldable}

\subsection{Knowledge distillation with limited data.}
\begin{foldable}
  The fundamental challenge of distillation on limited data is that the few available images cannot fully represent the actual images manifold.
  Consequently, the student fails to generalize to an unseen set.
  Thus, few-shot KD methods leverage data augmentation or generative models to augment the distillation set~\cite{kimura2018few,kong2020learning,wang2020neural,nguyen2022black}.
  Their approaches also vary depending on the accessibility of the teacher.
  FSKD~\cite{kimura2018few} uses trainable synthetic images that are part of the student's parameters.
  These images require adversarial signals from a white-box teacher for optimization.
  WaGe~\cite{kong2020learning} explores the data space in the approximate neighborhood of the few-shot data samples.
  This requires the logits of a white-box teacher for a Wasserstein-based loss.
  Most recently, BBKD~\cite{wang2020neural} and FS-BBT~\cite{nguyen2022black} only require a black-box teacher for the distillation.
  Both use MixUp to construct synthetic images for a larger distillation set for the student.
  While BBKD embraces active learning to choose synthetic images for distillation, FS-BBT trains a CVAE as an additional source of synthetic images.
  FS-BBT is the current SOTA few-shot KD method.
\end{foldable}

\subsection{Data-free knowledge distillation.}
\begin{foldable}
  As an even more extreme setting than few-shot, data-free KD methods assume no training data to train the student.
  To compensate, they often extract the knowledge from a white-box teacher \eg, its activation statistics~\cite{lopes2017data}, gradients~\cite{nayak2019zero,chen2019data}, and batch-norm layers~\cite{yin2020dreaming}.
  So far, only two methods use black-box teachers, namely ZSDB3KD~\cite{wang2021zero} and IDEAL~\cite{zhang2022ideal}.
  Although the data-free KD setting has several applications \eg, model inversion attack~\cite{zhang2020secret}, it is not really practical in the context of model compression.
  When we distill a large model, we often have data to validate the large and distilled models.
  Thus, few images can be also used for the distillation phase.
  As will be shown in \S\ref{ssec:04-experiments-comparison_zero_shot}., even with a small portion of data, our method significantly outperforms data-free KD methods.
\end{foldable}

\subsection{Generative Adversarial Networks in Knowledge Distillation}
\begin{foldable}
  Using Generative Adversarial Networks (GAN) to generate synthetic images is a popular approach in \textit{data-free} KD methods~\cite{chen2019data,addepalli2020degan,do2022momentum,zhang2022ideal}.
  However, our WGAN is trained with a novel scheme with \textit{three-fold differences}:
  \begin{enumerate}
    \item {
        \textbf{Generator:} Existing methods typically rely on the teacher's signals to tame the generator with extra losses \eg, one-hot, feature, and diversity losses.
        However, these losses are strictly inapplicable in our setting due to the absence of gradients from a black-box teacher.
      }
    \item {
        \textbf{Discriminator:} Existing methods cannot use a discriminator due to the lack of real images.
        In contrast, since our method allows to use any available real images, it uses the discriminator to distinguish real/fake images.
        Moreover, our method also trains the discriminator with high-confidence images to boost the diversity in image generation.
      }
    \item {
        \textbf{Teacher:} The teacher is trained for predicting image labels rather than distinguishing real/fake images.
        However, existing methods usually force the teacher to a discriminator's role, exposing it to knowledge gap.
        In contrast, we use teacher in a way where its intended strength is guaranteed to support our WGAN training.
      }
  \end{enumerate}
\end{foldable}

\section{Framework} \label{sec:03-framework}
\subsection{Problem statement}
\begin{foldable}
  Given a black-box pre-trained teacher network $T$ and a small set of labeled images $\mathcal{D}=\{x_{i},y_{i}\}_{i=1}^{N}$, our goal is to train a student network $S$ on $\mathcal{D}$ such that $S$ can approximate the performance of $T$.
\end{foldable}

\begin{foldable}
  A direct solution is to use standard KD~\cite{hinton2015distilling}, which trains $S$ to match both the ground-truth labels $y_{i}$ and the teacher's predictions $T(x_{i})$ via minimizing:
  \begin{equation}
    \mathcal{L}_\text{KD} = \mathbb{E}_{(x_{i}, y_{i}) \sim \mathcal{D}} \Bigl[{
        (1 - \lambda) \mathcal{L}_{\text{CE}}(S(x_{i}), y_{i}) +
        \lambda \mathcal{L}_{\text{KL}}(S(x_{i}), T(x_{i}))
    }\Bigr]
    \label{eq:StandardKD}
  \end{equation}
  where $\mathcal{L}_{\text{CE}}$ and $\mathcal{L}_{\text{KL}}$ are the cross-entropy and the Kullback-Leibler divergence losses, $S(x_{i})$ and $T(x_{i})$ are the student's and teacher's predictions (\ie, class probabilities), and $\lambda$ is a coefficient to balance the losses.
  Note that the \textit{temperature} factor in the original paper is not used as it requires access to the teacher's logits, which violates our black-box teacher assumption.
  Moreover, with a small distillation set $\mathcal{D}$, common KD approaches lead to a \textit{sub-optimal} model as they require lots of training images~\cite{hinton2015distilling,kim2018paraphrasing,ahn2019variational,tian2020contrastive}.
\end{foldable}

\begin{foldable}
  \textbf{Proposed Method.}
  To perform black-box few-shot KD, we propose our novel method \textbf{DivBFKD}.
  Our method has two phases: Generation and Distillation.
  In the Generation phase, we train a WGAN to produce synthetic images.
  In the Distillation phase, we train the student with the combined real and synthetic images.
\end{foldable}

\subsection{Generation phase.}
\begin{foldable}
  We train a WGAN model~\cite{gulrajani2017improved} with several key changes to generate diverse synthetic images.
  Like standard WGAN, our WGAN consists of a generator $G$ and a discriminator $D$.
  Iteratively, $G$ produces a synthetic image $\tilde{x} = G(z)$ from a random latent vector $z \sim \mathcal{N}(0, \mathbf{I})$.
  Meanwhile, $D$ predicts a scalar score indicating the likelihood of an input image being real.
  Jointly trained with opposite objectives, $G$ tries to fool $D$ by producing realistic images while $D$ tries to best distinguish between real images $x$ and synthetic images $\tilde{x}$.
  These objectives can be formulated into the generator loss $\mathcal{L}_G$ and the discriminator loss $\mathcal{L}_{D}$:
  \begin{equation}
    \mathcal{L}_G = -\mathbb{E}_{z \sim \mathcal{N}(0, \mathbf{I})} [D(G(z))],\label{eq:loss-G-wgan}
  \end{equation}
  \begin{equation}
    \mathcal{L}_{D} = \mathbb{E}_{z \sim \mathcal{N}(0, \mathbf{I})} [D(G(z))] - \mathbb{E}_{x \sim \mathcal{D}}[D(x)].\label{eq:loss-D-wgan}
  \end{equation}
\end{foldable}

\begin{foldable}
  However, if we only use the few-shot set $\mathcal{D}$ to train the WGAN, $G$ will generate synthetic images similar to the images in $\mathcal{D}$~\cite{karras2020training,yang2021data}, which are not diverse enough to facilitate effective distillation.
  To improve diversity of images generated by $G$, we propose the use of synthetic images that the teacher $T$ can predict confidently.
  We refer to these images as \textit{high-confidence} images.
\end{foldable}

\begin{foldable}
  \textbf{High-confidence images.}
  Given a synthetically generated (fake) image $\tilde{x}$, we define its \textit{confidence-score} $c_{\tilde{x}} = \max{k \in \{0, \ldots, K-1\}}T(\tilde{x})$, by taking the maximum over $T$'s predictive probability vector.
  For example, given three classes $\{0, 1, 2\}$ and class probabilities $T(\tilde{x})=[0.1, 0.7, 0.2]$, then $c_{\tilde{x}}=0.7$.
  We define $\tilde{x}$ a high-confidence image if $c_{\tilde{x}}$ is above a confidence-threshold $\tau\in[0,1]$.
  Intuitively, when $c_{\tilde{x}}$ is high, $\tilde{x}$ is likely close to the teacher's training images as it can be confidently classified.
\end{foldable}

\begin{figure}[h]
  \centering
  \includegraphics[viewport=0bp 20bp 455bp 174bp,clip,width=0.75\columnwidth]{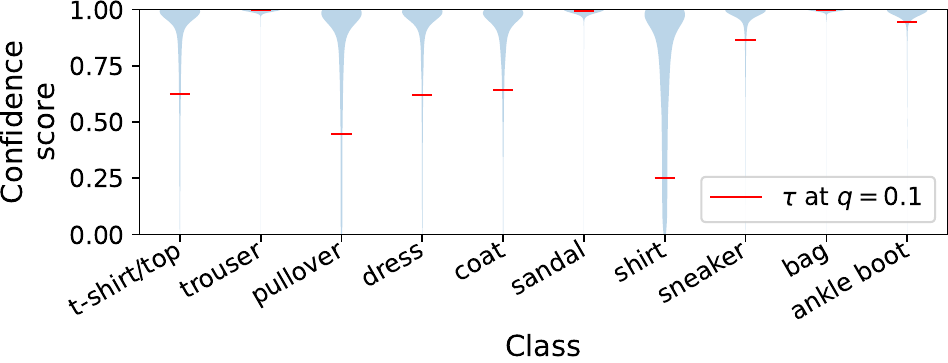}
  \caption{
    Distributions of confidence-scores across classes on FMNIST.
    They concentrate around 1 and have a long-tail.
    Red bars show adaptive confidence-thresholds $\tau$.
  }
  \label{fig:03-framework-pre_clamp_tau}
\end{figure}

\begin{foldable}
  \textbf{Adaptive thresholds.} The critical factor in determining a high-confidence image is the confidence-threshold $\tau$.
  We observe that the teacher has a class-specific bias, \ie, it predicts different classes with different confidence levels.
  Take the dataset FMNIST~\cite{xiao2017fashion} for example, the averaged confidence-score is $>0.99$ for classes `trouser' and `bag', but is only 0.75 for class `shirt' (Fig.~\ref{fig:03-framework-pre_clamp_tau}).
  In this case, fixing a single value $\tau=0.95$ may lead to high-confidence images being too lenient for `trouser' and `bags', but too strict for the class `shirt'.
  Thus, a single $\tau$ for all classes will make high-confidence images imbalanced.
  Further, considering any single class, the confidence-scores have a long-tailed distribution, where they concentrate near 1 and decay with lower values.
  Even though $\tau$ can be simply set to the averaged confidence-score, it would provide little flexibility and interpretability and also make high-confidence images biased towards \textit{outliers}.
\end{foldable}

\begin{foldable}
  To address these problems, we use adaptive thresholds $\{\tau^{k}\}_{k=0}^{K-1}$ that is class-specific based on the $q$-quantile of the confidence-scores of real images.
  Specifically, for each of the $K$ classes:
  \begin{equation}
    \tau^{k} = \text{quantile}_{q}(\{c_{x_{i}} \mid(x_{i}, y_{i}) \in \mathcal{D} \text{ and } y_{i} = k\}_{i=1}^{N}), q\in[0,1].
    \label{eq:adaptive-threshold}
  \end{equation}
  We choose $q = 0.1$, \ie, high-confidence images must have teacher's confidence higher than at least 10\% of those for the real ones, as perceived by the teacher $T$.
  This way, $\tau$ adaptively adjusts to any dataset and any class-wise bias from the teacher.
  Fig.~\ref{fig:03-framework-pre_clamp_tau} illustrates $\tau$ as red bars using $q = 0.1$ on FMNIST.
\end{foldable}

\begin{foldable}
  \textbf{Our WGAN training scheme.}
  We aim to harness the teacher's guidance to promote diversity of synthetic images.
  Fundamentally, synthetic images would be more diverse if the WGAN can learn on images that resemble the teacher's \textit{unknown} training images $\mathcal{D}^{*}$, whose diversity is superior to few-shot images $\mathcal{D}$.
  While there is no direct access to $\mathcal{D}^{*}$, high-confidence images can serve as a satisfactory alternative in terms of both quantity (new high-confidence images are generated at every WGAN training step) and quality (they likely reside close to $\mathcal{D}^{*}$, as we previously discussed).
  In other words, high-confidence images implicitly allow our WGAN to learn from $\mathcal{D}^{*}$.
\end{foldable}

\begin{figure}
  \centering
  \includegraphics[width=0.99\columnwidth]{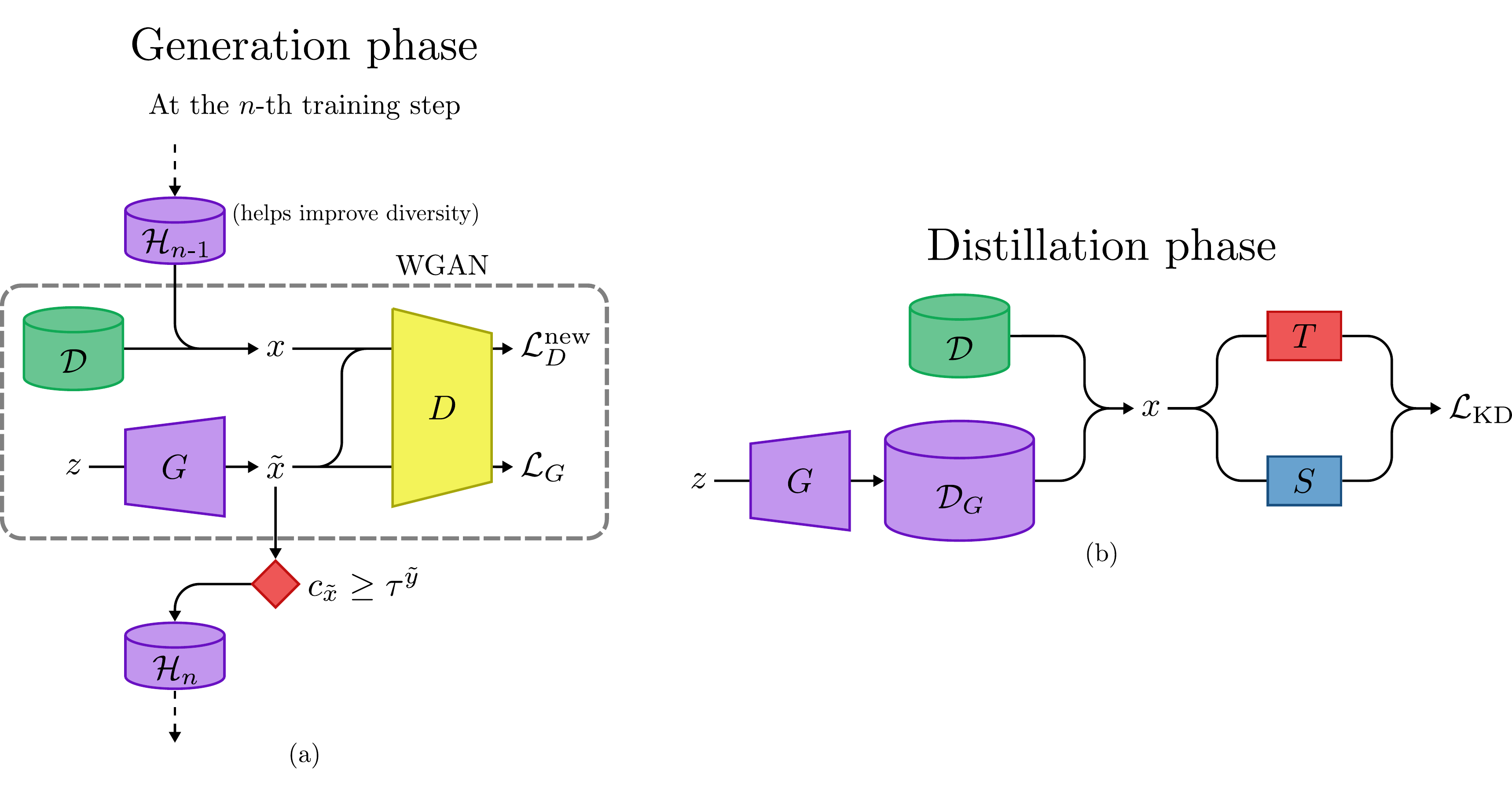}
  \caption{
    Illustration of our method DivBFKD.
    \textbf{(a)} \textbf{Generation:} We train our WGAN with the losses $\mathcal{L}_G$ and $\mathcal{L}_{D}^{\text{new}}$.
    When optimizing $G$, we construct the high-confidence set $\mathcal{H}_{n}=\{\tilde{x} = G(z) \mid c_{\tilde{x}} \geq \tau^{\tilde{y}}\}$ from synthetic images $\tilde{x}$, with their confidence-score $c_{\tilde{x}}$, pseudo-label $\tilde{y}$, and adaptive threshold $\tau^{\tilde{y}}$.
    When optimizing $D$, we sample $x$ from the combined real and high-confidence images set $\mathcal{D} \cup \mathcal{H}_{n-1}$.
    \textbf{(b) Distillation:} We augment few images in $\mathcal{D}$ with synthetic images in $\mathcal{D}_G$ to train the student $S$ on $\mathcal{D} \cup \mathcal{D}_G$.
  }
  \label{fig:03-framework-teaser}
\end{figure}

\begin{foldable}
  By this motivation, we introduce high-confidence images in the role of real images to the adversarial learning (Fig.~\ref{fig:03-framework-teaser}(a)), exposing our WGAN to a wider variety of training images.
  At each $n$-th training step:
  \begin{itemize}
    \item {
        The generator $G$ is trained with Eq~\ref{eq:loss-G-wgan}.
        For a fake image $\tilde{x}$, we compute its corresponding confidence-score $c_{\tilde{x}}$ and pseudo-label $\tilde{y} = \argmax_{k \in \{0, \ldots, K-1\}} T(\tilde{x})$.
        Using the adaptive thresholds $\{\tau^{k}\}_{k=0}^{K-1}$ computed with Eq~\ref{eq:adaptive-threshold}, we add qualified $\tilde{x}$ to the \textit{high-confidence} set $\mathcal{H}_{n} = \{\tilde{x} \mid c_{\tilde{x}} \geq \tau^{\tilde{y}}\}$.
        Then, $\mathcal{H}_{n}$ is stored to train the discriminator $D$ in the next step.
      }
    \item {
        Training the discriminator $D$ requires both real and fake images.
        However, our `real' images include real images that actually come from $\mathcal{D}$ and high-confidence images that come from the high-confidence set $\mathcal{H}_{n-1}$.
        Note that $\mathcal{H}_{n-1}$ is constructed in the previous step, and $\mathcal{H}_{0}=\emptyset$.
        With these changes, we train our discriminator with the new loss as:
      }
  \end{itemize}
  \begin{equation}
    \mathcal{L}_{D}^{\text{new}} = {
      \mathbb{E}_{z \sim \mathcal{N}(0, \mathbf{I})}[D(G(z))] -
      \mathbb{E}_{x \sim (\mathcal{D} \cup \mathcal{H}_{n-1})}[D(x)]
    }.
    \label{eq:loss-D-divbfkd}
  \end{equation}
\end{foldable}

\begin{foldable}
  Viewing as an adversarial game, when $D$ learns to appreciate high-confidence synthetic images as realistic, it also encourages $G$ to create more synthetic images like those.
  This interaction forms a positive feedback loop that helps $G$ to be more diverse.
  Furthermore, we emphasize that forming $\mathcal{H}_{n}$ is a \textit{cheap} operator and requires small memory.
  These modifications not only effectively improve the diversity of synthetic images, but also incur only trivial computation overhead.
\end{foldable}

\begin{foldable}
  \textbf{Summary of our novelty in WGAN.}
  We introduce a new training scheme for WGAN using high-confidence images based on the teacher-supervised adaptive thresholds.
  Since we aim to generate images close to the teacher's training images, our synthetic images are more \textit{diverse} and better cover \textit{unseen images}, significantly improving the student's generalization, as shown in \S\ref{sec:04-experiments}.
\end{foldable}

\begin{algorithm}[ht]
  \SetKwInOut{KwIn}{Input}
  \SetKwInOut{KwOut}{Output}
  \SetKwComment{Comment}{}{}

  \caption{Proposed method \textbf{DivBFKD}}
  \label{alg:algorithm-divbfkd}

  \KwIn{teacher $T$; labeled set $\mathcal{D} = \{x_{i}, y_{i}\}_{i=1}^{N}$; quantile $q$; synthetic budget $M$; batch size $B$}
  \KwOut{student network $S$}

  \vspace{-0.5\baselineskip}
  \Comment{\noindent\hrulefill}

  \vspace{0.5\baselineskip}
  Initialize generator $G$, discriminator $D$, student $S$, and high-confidence set $\mathcal{H}_0 \leftarrow \emptyset$ \\
  \Comment{$\triangleright$ \textbf{\textit{Stage 1: Generation}}}
  Infer $T$ on $\mathcal{D}$ to compute adaptive thresholds $\{\tau^k\}_{k=1}^K$ with $q$ \hfill Eq.~\ref{eq:adaptive-threshold} \\
  \While{$G$ not converged}{
    Sample latent vectors $\{z_i\}_{i=1}^B \sim \mathcal{N}(0,I)$ \\
    Generate synthetic images $\{\tilde{x}_i\}_{i=1}^B$ with $G$ \\
    Sample $\{x_i\}_{i=1}^B \sim (\mathcal{D} \cup \mathcal{H}_n)$ as real images \\
    Update $G$ on $\left\{\tilde{x}_i\right\}_{i=1}^B$ via $\mathcal{L}_G$ \hfill Eq.~\ref{eq:loss-G-wgan} \\
    Update $D$ on $\left\{x_i\right\}_{i=1}^B$ and $\left\{\tilde{x}_i\right\}_{i=1}^B$ via $\mathcal{L}_{D}^{new}$ \hfill Eq.~\ref{eq:loss-D-divbfkd} \\
    Compute pseudo-labels $\{\tilde{y}_i\}$ and confidence-scores $\{c_{\tilde{x}_i}\}$ for $\{\tilde{x}_i\}$ \\
    Construct $\mathcal{H}_n=\left\{\tilde{x_{i}} \mid c_{\tilde{x}_i}\ge\tau^{\tilde{y}_i}\right\}_{i=1}^B$ \\
  }

  \vspace{0.5\baselineskip}
  \Comment{$\triangleright$ \textbf{\textit{Stage 2: Distillation}}}
  Construct $\mathcal{D}_\text{KD}$ with $N$ real and $M$ synthetic images \hfill Eq.~\ref{eq:distill-set} \\
  \While{$S$ not converged}{
    Update $S$ on $\mathcal{D}_\text{KD}$ via $\mathcal{L}_\text{KD}$ \hfill Eq.~\ref{eq:loss-S}
  }

  \vspace{0.5\baselineskip}
  \Return{$S$}
\end{algorithm}

\subsection{Distillation phase}
\begin{foldable}
  After training WGAN in the Generation phase, we use it to generate synthetic images.
  We employ our generator $G$ to build a set of synthetic images $\mathcal{D}_G$, and their pseudo-labels (\ie, class probabilities) are obtained via the teacher $T$.
  Following BBKD~\cite{wang2020neural} and FS-BBT~\cite{nguyen2022black}, we construct a distillation set $\mathcal{D}_\text{KD}$ comprising of $N$ real images from $\mathcal{D}$ and $M$ synthetic images from $\mathcal{D}_G$:
  \begin{equation}
    \mathcal{D}_\text{KD} = {
      \{x_{i} \in \mathcal{D}\}_{i=1}^{N} \cup
      \{G(z_{j}) | z_{j} \sim \mathcal{N}(0, \mathbf{I})\}_{j=1}^{M}
    },
    \label{eq:distill-set}
  \end{equation}
  where $x_{i} \in \mathcal{D}$ and $\tilde{x}_{j} \in \mathcal{D}_G$ are real and synthetic images.
  Like other few-shot KD methods~\cite{wang2020neural,nguyen2022black}, we train the student $S$ with $\mathcal{D}_\text{KD}$ using a cross-entropy loss:
  \begin{equation}
    \mathcal{L}_\text{KD} = \mathbb{E}_{x_{i} \sim \mathcal{D}_\text{KD}} \mathcal{L}_\text{CE}(S(x_{i}), T(x_{i})).
    \label{eq:loss-S}
  \end{equation}
\end{foldable}

\begin{foldable}
  \textbf{Class balancing.} Inspired by~\cite{chen2019data,zhang2022ideal}, we generate a balanced set of synthetic images.
  Recalling that $M$ is the budget of synthetic images for KD and $K$ is the number of classes, each class would have $M / K$ synthetic images.
  We ensure class balance through \textit{rejection sampling} until we achieve the sufficient amount.
\end{foldable}

We summarize our method in Algorithm \ref{alg:algorithm-divbfkd}.

\section{Experiments} \label{sec:04-experiments}
\begin{foldable}
  We conduct a wide range of experiments to show our improvements over current SOTA few-shot KD baselines.
  We also include cross-architecture KD, visualization, ablation studies, and comparison with data-free KD methods.
  We provide more details in the \hyperref[sec:supp]{Supplementary}.
\end{foldable}

\subsection{Architectures and datasets}
\begin{foldable}
  We evaluate our method across seven benchmark image datasets: MNIST~\cite{lecun2002gradient}, FMNIST~\cite{xiao2017fashion}, SVHN~\cite{netzer2011reading}, CIFAR10, CIFAR100~\cite{krizhevsky2009learning}, Tiny-ImageNet~\cite{le2015tiny}, and Imagenette~\cite{howard2020imagenette}.
  These datasets cover different difficulty levels, including: (1) \textit{Simple} datasets with MNIST, FMNIST, SVHN, and CIFAR10 (10 classes) and (2) \textit{Complex} datasets with CIFAR100 (100 classes), Tiny-ImageNet (200 classes), and Imagenette (high-resolution at 224\texttimes 224).
\end{foldable}

\begin{foldable}
  We consider several popular network architectures for distillation, namely LeNet5~\cite{lecun2002gradient}, AlexNet~\cite{krizhevsky2012imagenet}, and ResNet~\cite{he2016deep}, and VGG~\cite{simonyan2014very}.
  These architectures and datasets are widely used to evaluate few-shot KD methods~\cite{kimura2018few,wang2020neural,nguyen2022black}.
\end{foldable}

\subsection{Baselines}
\begin{foldable}
  We compare our method DivBFKD with the following baselines:
  \begin{itemize}
    \item \textit{Student-Full}: The student is trained on the whole teacher's training set.
    \item \textit{Student-Alone}: The student is trained on the few-shot set $\mathcal{D}$.
    \item {
        \textit{Standard-KD}~\cite{hinton2015distilling}: The student is trained with the standard KD loss (Eq.~\ref{eq:StandardKD}) on $\mathcal{D}$.
        We select temperature $\lambda=0.9$, a standard value used in KD methods~\cite{hinton2015distilling,tian2020contrastive,yuan2020revisiting,ma2021undistillable}.
      }
    \item \textit{Few-shot KD} methods: We compare with two white-box methods, FSKD~\cite{kimura2018few} and WaGe~\cite{kong2020learning}, and two black-box methods, BBKD~\cite{wang2020neural} and FS-BBT~\cite{nguyen2022black}.
  \end{itemize}
\end{foldable}

\begin{foldable}
  For fair comparisons, we use the same teacher-student network architecture and the same numbers of real and synthetic images $N$ and $M$ as in other few-shot KD methods~\cite{kimura2018few,wang2020neural,nguyen2022black}.
  Their accuracy numbers are obtained from~\cite{nguyen2022black}\footnote{This is possible because we use benchmark datasets with fixed train and test splits.}.
  For our method, we set the quantile $q = 0.1$ to select high-confidence images across all experiments.
  We repeat each experiment three times with random seeds and report the averaged accuracy with standard deviation on the hold-out test set.
\end{foldable}

\subsection{Distillation performance}

\subsubsection{Simple datasets} \label{sec:experiments-simple-datasets}
\begin{table}[ht]
  \centering
  \begin{threeparttable}
    \caption{
      Classification accuracy (\%) on Simple datasets.
      $N$ and $M$ show the budget of real and synthetic images.
    }
    \label{tab:04-experiments-results_few_classes}
    \begin{tabular}{l l l l l}
      \toprule
      \textbf{Dataset} & \textbf{MNIST} & \textbf{FMNIST} & \textbf{SVHN} & \textbf{CIFAR10} \\
      $(N,M)$ & (2\,K, 24\,K) & (2\,K, 48\,K) & (2\,K, 40\,K) & (2\,K, 40\,K) \\
      \midrule
      Teacher                   & 99.28                   & 90.90                   & 96.16                   & 90.05 \\
      Student-Full              & \meanstd{98.91}{0.08}   & \meanstd{88.68}{0.53}   & \meanstd{95.61}{0.07}   & \meanstd{87.58}{0.36} \\
      \midrule

      \multicolumn{5}{c}{\textbf{Smaller Architecture}} \\
      \midrule
      Student-Alone             & \meanstd{95.14}{0.19}   & \meanstd{79.24}{1.05}   & \meanstd{85.68}{0.46}   & \meanstd{59.02}{0.54} \\
      Standard-KD\blackbox      & \meanstd{95.36}{0.11}   & \meanstd{80.76}{0.37}   & \meanstd{89.11}{0.22}   & \meanstd{59.77}{0.96} \\
      FSKD\whitebox             & 80.43\dg                & 68.64\dg                & $-$                     & 40.58\dg \\
      WaGe\whitebox             & $-$                     & 85.18\dg                & $-$                     & 73.08\dg \\
      FS-BBT\blackbox           & 98.42\dg                & 84.73\dg                & $-$                     & 74.10\dg \\
      \textbf{DivBFKD}\blackbox & \meanstdbf{98.58}{0.06} & \meanstdbf{86.50}{0.18} & \meanstdbf{94.47}{0.12} & \meanstdbf{76.97}{0.56} \\
      \midrule

      \multicolumn{5}{c}{\textbf{Same Architecture}} \\
      \midrule
      BBKD\blackbox             & 98.74\dg                & 80.90\dg                & $-$                     & 74.60\dg \\
      FS-BBT\blackbox           & 98.91\dg                & 86.53\dg                & $-$                     & 76.17\dg \\
      \textbf{DivBFKD}\blackbox & \meanstdbf{98.97}{0.04} & \meanstdbf{87.65}{0.37} & \meanstdbf{94.61}{0.08} & \meanstdbf{78.07}{0.26} \\
      \bottomrule
    \end{tabular}
    \begin{tablenotes}[flushleft] \footnotesize
    \item {
        [(W)] white-box; [(B)] black-box; [$-$] not reported; [$\dagger$] reported from respective paper; [\textbf{bold}] best results.
      }
    \end{tablenotes}
  \end{threeparttable}
\end{table}

\begin{foldable}
  Following~\cite{nguyen2022black}, we conduct distillation in two settings:
  \begin{enumerate}
    \item \textit{Smaller} architecture: The student has a smaller architecture than the teacher: LeNet5 to LeNet5-Half on MNIST, FMNIST; and AlexNet to AlexNet-Half on SVHN, CIFAR10.\footnote{The \textit{-Half} version has half the channels in each layer.}
    \item \textit{Same} architecture: The student has an identical architecture to the teacher.
  \end{enumerate}
\end{foldable}

\begin{foldable}
  We report the results in Table~\ref{tab:04-experiments-results_few_classes} and make the following observations:
  \begin{itemize}
    \item Our method DivBFKD improves by around 3-6\% over Standard-KD on MNIST, FMNIST, and SVHN and by significantly 17\% on CIFAR10.
    \item {
        Compared to other few-shot KD methods, our DivBFKD also excels in both smaller and same architecture settings.
        It improves upon second-best FS-BBT by around 2\% on FMNIST and CIFAR10.
      }
    \item {
        The performance gap between our DivBFKD and its upper bound Student-Full is very close on MNIST (0.33\%), FMNIST (2.18\%), and SVHN (1.14\%).
        We emphasize that Student-Full uses the teacher's training set ($\ge$ 60\,K real images), much larger than the 2\,K real images we started with.
      }
  \end{itemize}
\end{foldable}

\subsubsection{Complex datasets} \label{sec:experiments-complex-datasets}
\begin{foldable}
  Similar to~\cite{wang2020neural,nguyen2022black,kong2020learning}, we distill from ResNet32 to ResNet20 on CIFAR100 and Tiny-ImageNet (Table~\ref{tab:ResultsHundredClasses}).
  With limited data, Student-Alone loses roughly half of its accuracy compared to Student-Full.
  As a naive solution, Standard-KD improves over Student-Alone 10\% on CIFAR100 and 14\% on Tiny-ImageNet.
  Ultimately, our method DivBFKD is the best method, exceeding the second-best FS-BBT by 3\% on CIFAR100 and 1\% on Tiny-ImageNet.
\end{foldable}

\begin{foldable}
  We further evaluate our method on high-resolution images with Imagenette.
  On 2\,K real images, Student-Alone achieves 76.80\% while Standard-KD marginally improves to 76.94\%.
  DivBFKD further improves upon this result by 10\%, approximating Student-Full.
\end{foldable}

\begin{table}[ht]
  \centering
  \begin{threeparttable}
    \caption{
      Classification accuracy (\%) on Complex datasets.
      $N$ and $M$ show the budget of real and synthetic images.
    }
    \label{tab:ResultsHundredClasses}
    \begin{tabular}{l l l l}
      \toprule
      \textbf{Dataset} & \textbf{CIFAR100} & \textbf{Tiny-ImageNet} & \textbf{Imagenette} \\
      $(N,M)$                   & (5\,K, 40\,K)           & (10\,K, 50\,K)        & (2\,K, 40\,K) \\
      \midrule
      Teacher                   & 71.41                   & 56.27                 & 90.47 \\
      Student-Full              & \meanstd{68.28}{0.20}   & \meanstd{51.51}{0.31} & \meanstd{89.89}{0.41} \\
      \midrule
      Student-Alone             & \meanstd{35.71}{0.39}   & \meanstd{26.59}{0.25} & \meanstd{76.80}{0.78} \\
      Standard-KD\blackbox      & \meanstd{45.76}{0.48}   & \meanstd{40.42}{0.19} & \meanstd{76.94}{0.14} \\
      WaGe\whitebox             & 20.32\dg                & $-$                   & $-$ \\
      BBKD\blackbox             & 53.41\dg                & 40.01\dg              & $-$ \\
      FS-BBT\blackbox           & 56.28\dg                & 43.29\dg              & $-$ \\
      \textbf{DivBFKD}\blackbox & \meanstdbf{59.91}{0.31} & \meanstdbf{44.25}{0.38} & \meanstdbf{86.48}{0.72} \\
      \bottomrule
    \end{tabular}
    \begin{tablenotes}[flushleft] \footnotesize
    \item[] {
        [(W)] white-box; [(B)] black-box; [$-$] not reported; [$\dagger$] reported from respective paper; [\textbf{bold}] best results.
      }
    \end{tablenotes}
  \end{threeparttable}
\end{table}

\subsubsection{Cross-architecture distillation} \label{sec:experiments-crossarch}
\begin{foldable}
  In previous subsections, we distill from the teacher to a student networks of similar architectures to have a consistent setting with other few-shot KD methods (\eg, LeNet5 to LeNet5-Half, AlexNet to AlexNet-Half, and ResNet32 to ResNet20).
  This section further explores the cases where the teacher and student architectures belong to different families: ResNet~\cite{he2016deep}, VGG~\cite{simonyan2014very}, and AlexNet~\cite{krizhevsky2012imagenet}.
  We adopt the same setting from the standard experiment on CIFAR10 from \S\ref{sec:experiments-complex-datasets}.
  Table~\ref{tab:Cross-architecture} shows the performance of our method DivBFKD with three teachers ResNet32, VGG16, and AlexNet combined with three students ResNet20, VGG11, and AlexNet-Half on CIFAR10.
\end{foldable}

\begin{table}[ht]
  \caption{Student accuracy (\%) of cross-architecture distillation on CIFAR10.}
  \label{tab:Cross-architecture}
  \centering
  \begin{threeparttable}
    \begin{tabular}{c c|c c c c c}
      \toprule
      \multicolumn{2}{c|}{\textbf{Architecture}} & \multicolumn{5}{c}{\textbf{Accuracy}}\\
      Teacher & Student & Teacher & S-Full & S-Alone & Standard-KD & DivBFKD \\
      \midrule
      ResNet32 & ResNet20     & 93.34 & & & \meanstd{65.86}{1.90} & \meanstd{84.05}{0.10} \\
      VGG16    & ResNet20     & 91.30 & \meanstd{92.34}{0.06} & \meanstd{64.52}{1.60} & \meanstd{64.76}{0.89} & \meanstd{80.69}{0.48} \\
      AlexNet  & ResNet20     & 90.05 & & & \meanstd{65.29}{0.89} & \meanstd{79.57}{0.16} \\
      \midrule
      ResNet32 & VGG11        & 93.34 & & & \meanstd{53.78}{0.78} & \meanstd{69.42}{0.83} \\
      VGG16    & VGG11        & 91.30 & \meanstd{89.12}{0.17} & \meanstd{53.34}{0.57} & \meanstd{53.80}{1.21} & \meanstd{75.41}{0.06} \\
      AlexNet  & VGG11        & 90.05 & & & \meanstd{53.51}{0.14} & \meanstd{74.60}{0.29} \\
      \midrule
      ResNet32 & AlexNet-Half & 93.34 & & & \meanstd{60.36}{0.82} & \meanstd{74.70}{0.31} \\
      VGG16    & AlexNet-Half & 91.30 & \meanstd{87.58}{0.36} & \meanstd{59.02}{0.54} & \meanstd{60.47}{0.10} & \meanstd{76.16}{0.57} \\
      AlexNet  & AlexNet-Half & 90.05 & & & \meanstd{59.77}{0.96} & \meanstd{76.97}{0.56} \\
      \bottomrule
    \end{tabular}
    \begin{tablenotes}[flushleft] \footnotesize
    \item {Same teacher-student pairs share the same Student-Full and Student-Alone accuracy.}
    \end{tablenotes}
  \end{threeparttable}
\end{table}

\begin{foldable}
  We have the following observations:
  \begin{itemize}
    \item Among three architecture families, ResNet is the best one with a 93.34\% Teacher, 92.34\% Student-Full, and 64.52\% Student-Alone.
    \item Standard-KD is helpful and it is always better than Student-Alone in all teacher-student combinations.
    \item {
        VGG is better than AlexNet when the teacher and the student are trained on the whole training set.
        However, VGG student is worse than AlexNet student when it is trained on the few-shot dataset.
        Namely, Student-Alone of VGG11 has 53.34\% accuracy while Student-Alone of AlexNet-Half has 59.02\% accuracy.
      }
    \item Our method DivBFKD remarkably improves accuracy over Standard-KD by 14--22\% on all teacher-student combinations.
    \item Distillation is often most effective when the teacher's and student's architecture belong to the same family.
  \end{itemize}
  Overall, the robust performance across multiple data scales and architectures suggests that our method successfully tackles the black-box few-shot KD setting.
\end{foldable}

\subsection{Diversity of our synthetic images}

\subsubsection{Comparison with the teacher's training images.}

\begin{figure}[ht]
  \centering
  \includegraphics[width=1\columnwidth]{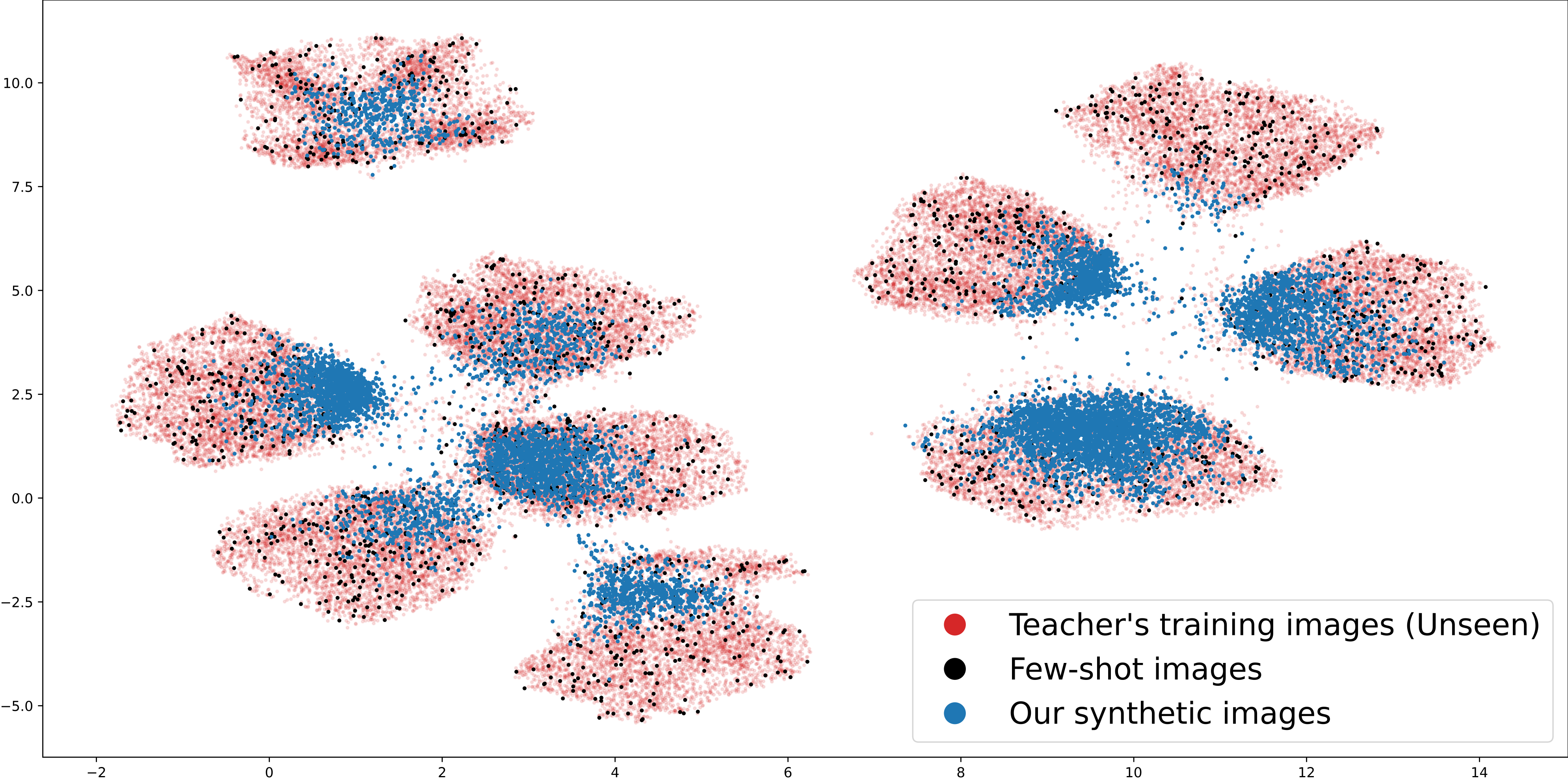}
  \caption{Embeddings of teacher's training images, few-shot images, and our synthetic images on CIFAR10.}
  \label{fig:Distributions-of-CIFAR10}
\end{figure}

\begin{foldable}
  As we mentioned earlier, our goal is to generate synthetic images close to the teacher's training images.
  To prove this argument, we embed and visualize the teacher's training images (they are \textit{unknown} and not used in the KD process) and our synthetic images of CIFAR10 in Fig.~\ref{fig:Distributions-of-CIFAR10}.
  It shows that our synthetic images belong to the distribution of the teacher's training images.
  With the teacher's supervision, our synthetic images show their strong resemblance to the teacher's training images.
\end{foldable}

\begin{foldable}
  For quantitative results, we compute the \textit{Coverage} metric~\cite{alaa2022faithful} to measure the fraction of teacher's training samples whose neighborhoods contain at least one synthetic sample.
  Our synthetic images achieve 0.42, compared to 0.18 of the few-shot images and 0.29 of the synthetic images generated from the standard WGAN.
  This shows that our synthetic images resemble many variability of the teacher's training images, thus improving image diversity.
\end{foldable}

\begin{figure}[ht]
  \centering
  \begin{subfigure}[t]{0.475\columnwidth}
    \centering
    \includegraphics[width=\linewidth]{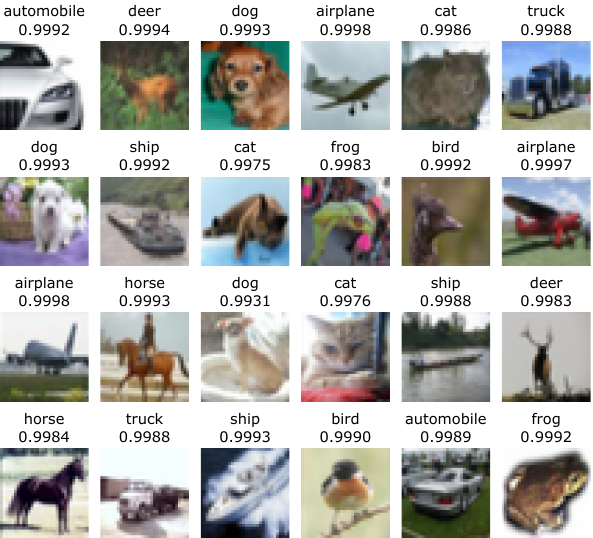}
    \caption{Real images}
  \end{subfigure}
  \hfill
  \begin{subfigure}[t]{0.475\columnwidth}
    \centering
    \includegraphics[width=\linewidth]{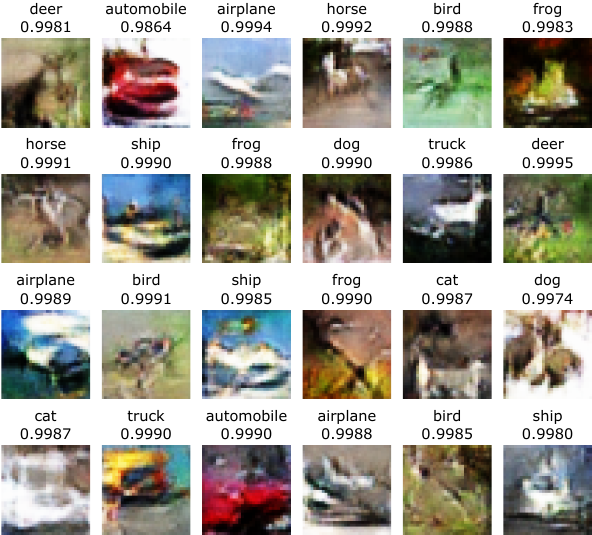}
    \caption{Synthetic images}
  \end{subfigure}
  \caption{Visualization of real and synthetic images along with their predictive labels and confidence-scores provided by the teacher on CIFAR10.}
  \label{fig:Visualization}
\end{figure}

\subsubsection{Quality and quantitative measurements}
\begin{foldable}
  We randomly visualize 24 real images and 24 synthetic images generated by our WGAN in Fig.~\ref{fig:Visualization}.
  Primarily, our goal is not to generate visually aesthetic images but to generate diverse images that the teacher confidently predicts their labels.
  As a side note, image quality is also not a strict requirement in KD methods relied on synthetic images~\cite{chen2019data,wang2020neural,wang2021zero}.
  However, our synthetic images in Fig.~\ref{fig:Visualization}(b) are able to show convincing properties of their real counterparts.
\end{foldable}

\begin{foldable}
  \textbf{Inception score (IS) and Fr\'{e}chet inception distance (FID).} To quantify the diversity and quality of synthetic images, we compute the IS~\cite{salimans2016improved} (higher is better) and FID~\cite{heusel2017gans} (lower is better) of our synthetic images in Table~\ref{tab:IS-FID}.
  We compare with the standard WGAN and FS-BBT (the second-best baseline), where our method achieves the best scores.
  These results confirm that our method generate synthetic images with high diversity and quality, and explain their effectiveness for the distillation step.
\end{foldable}

\begin{table}[ht]
  \centering
  \begin{threeparttable}
    \caption{IS and FID of synthetic images on CIFAR10.}
    \label{tab:IS-FID}
    \begin{tabular}{l c c}
      \toprule
      \textbf{Method} & \textbf{IS ($\uparrow$)} & \textbf{FID ($\downarrow$)} \\
      \midrule
      CVAE (from~\cite{nguyen2022black}) & 3.51          & 19.91 \\
      Standard WGAN                      & 3.99          & 14.79 \\
      WGAN with AT (Ours)                & \textbf{4.42} & \textbf{13.96} \\
      \bottomrule
    \end{tabular}
    \begin{tablenotes}[flushleft] \footnotesize
    \item {[AT] adaptive thresholds.}
    \end{tablenotes}
  \end{threeparttable}
\end{table}

\subsection{Ablation studies}

\subsubsection{Impacts of different components}
\begin{foldable}

  As described in \S\ref{sec:03-framework}, the core components of our method include WGAN and adaptive thresholds in the generation phase, and class balancing in the distillation phase.
  Adding both latter components to WGAN completes our method.
  Table~\ref{tab:Ablation-study-Components} breaks down the contribution of each component, where both adaptive thresholds (AT) and class balancing (CB) improve our accuracy when applied either independently or jointly.
\end{foldable}

\begin{table}[ht]
  \centering
  \begin{threeparttable}
    \caption{Student accuracy (\%) with incremental components of DivBFKD.}
    \label{tab:Ablation-study-Components}
    \begin{tabular}{l c c}
      \toprule
      \textbf{Components}                 & \textbf{CIFAR10}        & \textbf{Tiny-ImageNet} \\
      \midrule
      WGAN~\cite{arjovsky2017wasserstein} & \meanstd{75.59}{0.69}   & \meanstd{42.70}{0.25} \\
      WGAN with AT                        & \meanstd{76.35}{0.21}   & \meanstd{43.25}{0.71} \\
      WGAN with CB                        & \meanstd{76.88}{0.45}   & \meanstd{43.75}{0.40} \\
      \textbf{DivBFKD}                    & \meanstdbf{76.97}{0.56} & \meanstdbf{44.25}{0.38} \\
      \bottomrule
    \end{tabular}
    \begin{tablenotes}[flushleft] \footnotesize
    \item {[AT] adaptive thresholds; [CB] class balancing.}
    \end{tablenotes}

  \end{threeparttable}
\end{table}

\subsubsection{Impacts of the numbers of real and synthetic images}
\begin{figure}[ht]
  \centering
  \includegraphics[width=0.485\columnwidth]{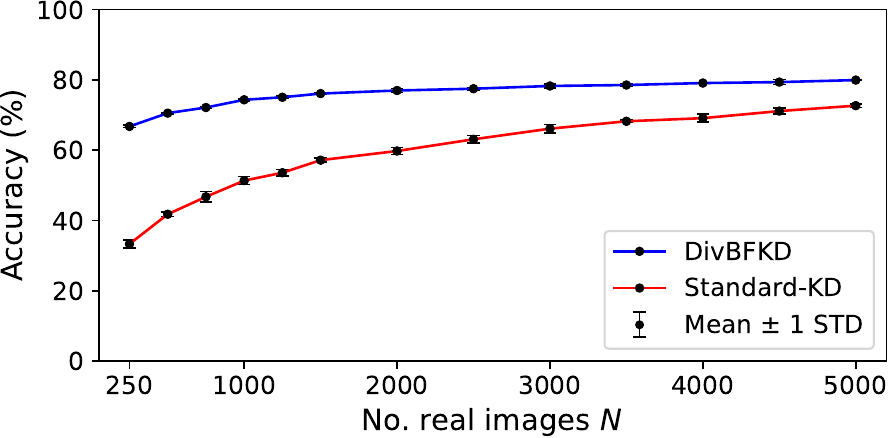}\hspace{0.2cm}\includegraphics[width=0.485\columnwidth]{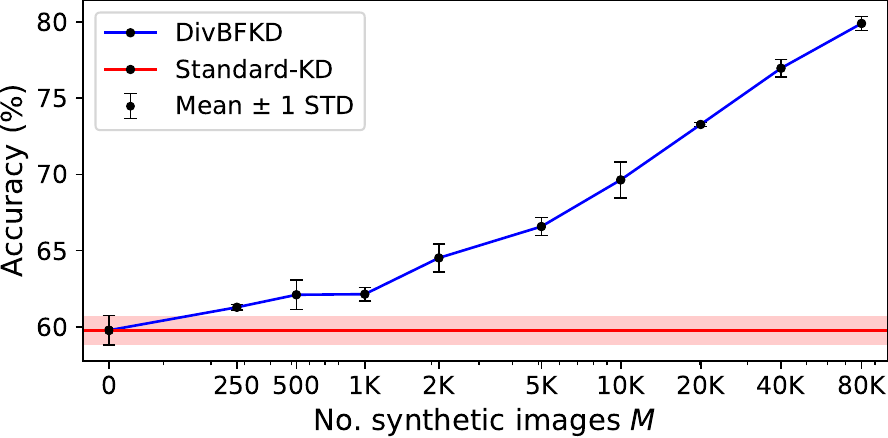}
  (a)\hspace{0.49\columnwidth}(b)
  \caption{Accuracy versus the budget of real ($N$) and synthetic images ($M$) on CIFAR10.}
  \label{fig:04-experiments-abla_data}
\end{figure}

\begin{foldable}
  We investigate our performance with varying amounts of data on CIFAR10.
  Fig.~\ref{fig:04-experiments-abla_data}(a) shows that both Standard-KD and our DivBFKD improve student accuracy with more real images, as expected.
  At $N$ = 5\,K, our DivBFKD achieves 79.93\%, while Standard-KD only achieves 72.66\%.
  Conversely, at $N$ = 250, Standard-KD drops abruptly to 33.33\%, while our method still has an acceptable 66.76\%.
  This trend proves that our method can still distill robustly in extremely limited data cases.
  Besides, we also expect that our performance should improve with more synthetic images.
  This agrees with Fig.~\ref{fig:04-experiments-abla_data}(b) where our DivBFKD clearly outperforms Standard-KD and improves with larger $M$.
\end{foldable}

\subsubsection{Impact of the quantile}
\begin{figure}[ht]
  \centering
  \includegraphics[width=0.75\columnwidth]{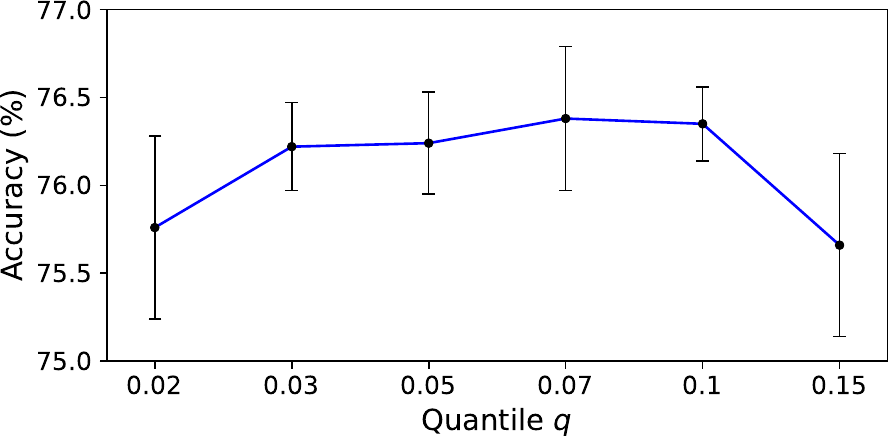}
  \caption{Accuracy versus the quantile $q$ on CIFAR10.}
  \label{fig:Results-Hyperparameter-tuning}
\end{figure}

\begin{foldable}
  In our method, the quantile $q$ is principally used to compute adaptive thresholds to determine high-confidence images.
  Contrarily, while too small $q$ might introduce counter-productive noise, too large $q$ does not allow enough high-confidence images to improve diversity.
  We achieve a relatively stable accuracy of 76.22--76.97\% with $q\in[0.03,0.10]$ (Fig.~\ref{fig:Results-Hyperparameter-tuning}) and recommend this range for a good trade-off between image diversity and training stability.
\end{foldable}

\subsection{Comparison with data-free KD methods} \label{ssec:04-experiments-comparison_zero_shot}
\begin{foldable}
  We also compare our method with several popular data-free (zero-shot) KD methods, including four white-box methods: Meta-KD~\cite{lopes2017data}, ZSKD~\cite{nayak2019zero}, DAFL~\cite{chen2019data}, and DFKD~\cite{wang2021data}, and two black-box methods ZSDB3KD~\cite{wang2021zero} and IDEAL~\cite{zhang2022ideal}.
  From Table~\ref{tab:04-experiments-comparison_data_free}, our method DivBFKD can successfully take advantage of a small amount of data (2\,K images) and perform much better than data-free methods on FMNIST and CIFAR10, while it is comparable on MNIST.
\end{foldable}

\begin{table}[ht]
  \centering
  \begin{threeparttable}
    \caption{Classification accuracy (\%) comparison with zero-shot KD methods.}
    \label{tab:04-experiments-comparison_data_free}
    \begin{tabular}{l|c|c|c}
      \hline
      \textbf{Dataset} & \textbf{MNIST} & \textbf{FMNIST} & \textbf{CIFAR10}  \\
      \hline
      Meta-KD\whitebox          & 92.47          & $-$            & $-$ \\
      ZSKD\whitebox             & 98.77          & 79.62          & 69.56 \\
      DAFL\whitebox             & 98.20          & $-$            & 66.38 \\
      DFKD\whitebox             & \textbf{99.08} & $-$            & 73.91 \\
      ZSDB3KD\blackbox          & 96.54          & 72.31          & 59.46 \\
      IDEAL\blackbox            & 96.32          & 83.92          & 65.61 \\
      \textbf{DivBFKD}\blackbox & 98.97          & \textbf{87.65} & \textbf{78.07} \\
      \hline
    \end{tabular}
    \begin{tablenotes}[flushleft] \footnotesize
    \item {[(W)] white-box; [(B)] black-box; [$-$] not reported; [\textbf{bold}] best results.}
    \end{tablenotes}
  \end{threeparttable}
\end{table}

\section{Conclusion} \label{sec:05-conclusion}
While a large training set and internal access to the teacher are the prerequisites to most knowledge distillation (KD) approaches, they are rarely available due to various restrictions.
To address these practical challenges, we propose DivBFKD, a novel method to bridge the gap for this black-box few-shot KD problem.
We investigate the diversity of training data, a crucial but inadequately addressed factor for KD.
Specifically, we propose the use of high-confidence images---ones that the teacher can classify confidently.
We also define adaptive thresholds, the teacher-supervised criteria to negate teacher's class-wise bias during the selection of these images.
Under the teacher's supervision, we introduce high-confidence images to the WGAN training loop on-the-fly to boost the diversity of image generation process with negligible cost.
Through comprehensive evaluation, we have shown that our DivBFKD could leverage diversity to achieve state-of-the-art KD performance among other few-shot KD methods on multiple image datasets and architectures at different scales.
As a future work, it would be interesting to study the robustness aspects of the distilled model relative to the original model to have the assurance~\cite{gopakumar2018algorithmic} for successful deployment in the real world.

\bibliography{references}

\appendix
\newpage
\label{sec:supp}
\begin{center}
    {\Large [Supplementary Material]} \\[1.5em] 
    {\LARGE Improving Diversity in \\[0.5em] Black-box Few-shot Knowledge Distillation} \\[1.5em]
  \end{center}
\runningheads{T-N. Vo et al.}{Supplementary Material: DivBFKD}
\begin{foldable}
  In this supplementary, we provide the implementation details on our experiments, including description on the datasets (full and few-shot) as well as the architecture and training of our teacher, student, and WGAN networks.
\end{foldable}

\section{Datasets} \label{sec:supp-datasets}
\begin{table}[ht]
  \centering
  \begin{threeparttable}
    \caption{
      Dataset details and their augmentation transforms.
      $N$ ($\times 10^3$) denotes the number of images in thousands and $K$ denotes the number of classes.
      Dimension (images and transforms) are in pixels (px).
    }
    \label{tab:Dataset-details}
    \begin{tabular}{c c c c c}
      \toprule
      \textbf{Dataset} & $N~(\times 10^3)$ & $K$ & \textbf{Dimension} & \textbf{Transforms} \\
      \midrule
      MNIST         & 60  & 10   & $1\times 28 \times 28$          & Resize to 32$\times$32 \\
      FMNIST       & 60  & 10   & $1\times 28 \times 28$           & Resize to 32$\times$32 \\
      SVHN          & 73  & 10   & $3\times 32 \times 32$          & $\Bigl[$ \makecell{Zero-padding (4 px/side) \\ Random crop to $32\times32$} $\Bigr]$\\
      CIFAR10       & 50  & 10   & $3\times 32 \times 32$          & $\Biggl[$ \makecell{Zero-padding (4 px/side) \\ Random horizontal flip \\ Random crop to $32\times32$} $\Biggr]$\\
      CIFAR100      & 50  & 100  & $3\times 32 \times 32$          & $\Biggl[$ \makecell{Zero-padding (4 px/side) \\ Random horizontal flip \\ Random crop to $32\times32$} $\Biggr]$\\
      Tiny-ImageNet & 100 & 200  & $3\times 64 \times 64$          & $\Biggl[$ \makecell{Zero-padding (8 px/side) \\ Random horizontal flip \\ Random crop to $64\times64$} $\Biggr]$\\
      Imagenette    & 9.5 & 10   & $3\times 408 \times 472 ^{(*)}$ & $\Biggl[$ \makecell{Resize to $256\times256$ \\ Random horizontal flip \\ Random crop to $224\times224$} $\Biggr]$\\
      \bottomrule
    \end{tabular}
    \begin{tablenotes}[flushleft] \footnotesize
    \item {[$*$] Imagenette dimensions are reported as mean values}
    \end{tablenotes}
  \end{threeparttable}
\end{table}

\begin{foldable}
  We use a total of seven image classification datasets for our experiments: MNIST~\cite{lecun2002gradient}, FMNIST~\cite{xiao2017fashion}, SVHN~\cite{netzer2011reading}, CIFAR10, CIFAR100~\cite{krizhevsky2009learning}, Tiny-ImageNet~\cite{le2015tiny}, and Imagenette~\cite{howard2020imagenette}.
  Only MNIST and FMNIST have black-and-white images (1 color channel) while the other datasets have colored images (3 color channels).
  Depending on their relative difficulty, we only use basic image transformations to augment our data, including resizing, zero-padding, random horizontally flipping, random cropping.
  We present the datasets' details and corresponding transforms in Table~\ref{tab:Dataset-details}.
  For Imagenette, we report the average value; their image dimension ranges from 27$\times$80 to 4368$\times$2912.
\end{foldable}

\section{Construction of the few-shot dataset} \label{sec:supp-construction-fewshot}
\begin{foldable}
  Following~\cite{wang2020neural,nguyen2022black}, to construct the few-shot dataset $\mathcal{D}$, we used $N = 2000$ images for MNIST, FMNIST, SVHN, CIFAR10, and Imagenette, $N=5000$ images for CIFAR100, and $N=10000$ images for Tiny-ImageNet.
  They were randomly drawn from the original training set such that each class had $N/K$ images ($K$ is the number of classes).
\end{foldable}

\section{Training setting of the Teacher} \label{sec:supp-training}
\begin{foldable}
  We trained the teacher network with a stochastic gradient descent (SGD) optimizer using 0.9 momentum, $5\times10^{-4}$ weight decay, and 128 batch size.
  We fixed the learning rate to $10^{-2}$ on MNIST and FMNIST.
  On other datasets, we scheduled the learning rate to $10^{-1}$ at the beginning and decreased it by ten times at 50\% and 75\% progress.
  We trained for 100 epochs on MNIST and FMNIST, and 200 epochs on other datasets.
\end{foldable}

\section{Our WGAN architecture} \label{sec:supp-wgan-arch}
\begin{foldable}
  We employed a deep convolutional GAN as in~\cite{radford2015unsupervised}, where the generator and discriminator had symmetrical architectures:
  \begin{itemize}
    \item {
        The generator accepts latent vectors of dimension 100 for MNIST, 200 for FMNIST, and 256 for other datasets.
        The latent inputs are then projected to 256 base feature maps of dimension 8$\times$8 (except for Imagenette, 7$\times$7), which then undergo batch normalization.
        These feature maps are passed through upscaling convolutional blocks, each consisting of a nearest neighbor (2$\times$) upsample layer, a convolutional layer with half the channels, a batch normalization, and a leaky ReLU activation.
        Finally, once the feature maps have identical dimension to that of the training images, they are compressed to either grayscale or RGB with a final convolutional layer and a sigmoid activation to map the output pixel values to range $[0, 1]$.
      }
    \item {
        The discriminator feeds forward input images through convolutional blocks, each consisting of a (0.5$\times$) downscaling convolutional layer, a batch normalization (except for the first block), and a leaky ReLU activation.
        While the final block has 256 base feature maps of dimension 8$\times$8 (except for Imagenette, 7$\times$7), each previously adjacent block has half the channels.
        These base feature maps are eventually passed through a linear layer to yield a scalar score.
      }
  \end{itemize}
\end{foldable}

\section{Training setting of our WGAN} \label{sec:supp-training-wgan}
\begin{foldable}
  We followed the implementation in~\cite{gulrajani2017improved} to train our WGAN~\cite{arjovsky2017wasserstein}.
  As recommended, we enforced the Lipschitz constraint on the discriminator with gradient penalty, whose coefficient with the Wasserstein loss was set to 10:1.
  The number of discriminator updates per generator updates was set to 5.
  However, we alternatively used the RMSProp optimizer with learning rate $5 \times 10^{-5}$ and batch size 250 for both models.
  We used the number of epochs of 500 for MNIST, 1000 for FMNIST and SVHN, 2000 for CIFAR10 and Imagenette, and 5000 for CIFAR100 and Tiny-ImageNet.
\end{foldable}

\section{Training setting of our Student} \label{sec:supp-training-student}
\begin{foldable}
  For a fair comparison, we used the same number of synthetic images for distillation as in other few-shot KD methods~\cite{kimura2018few,wang2020neural,nguyen2022black}, which is 24\,K for MNIST, 48\,K for FMNIST, 40\,K for SVHN, CIFAR10, CIFAR100, and Imagenette, and 50\,K for Tiny-ImageNet.
  We trained the Student with an SGD optimizer using 0.9 momentum and $5 \times 10^{-4}$ weight decay.
  We fixed the learning rate to $10^{-2}$ on MNIST and FMNIST.
  On other datasets, we scheduled the learning rate to $10^{-1}$ at the beginning and decrease it by ten times at 50\% and 75\% distillation progress.
  We used a batch size of 250 and the number of epochs of 100 for MNIST, 200 for FMNIST and SVHN, 400 for CIFAR10 and Imagenette, and 1000 for CIFAR100 and Tiny-ImageNet.
\end{foldable}

\end{document}